\documentclass{article}

    \usepackage[nonatbib, preprint]{neurips_2019}

\usepackage[utf8]{inputenc} 
\usepackage[T1]{fontenc}    
\usepackage{hyperref}       
\usepackage{url}            
\usepackage{booktabs}       
\usepackage{amsfonts}       
\usepackage{nicefrac}       
\usepackage{microtype}      

\usepackage[
backend=biber,
style=numeric,
]{biblatex}
\usepackage{graphicx} 
\graphicspath{ {./image/} }

\usepackage[rightcaption]{sidecap}

\usepackage{wrapfig}

\title{Innovative Drug-like Molecule Generation from Flow-based Generative Model}

\author{%
  Haotian Zhang \\
  School of Medicine \\
  University of Pittsburgh\\
  Pittsburgh, PA 15213 \\
  \texttt{haotian3@andrew.cmu.edu} \\
  \And
  Linxiaoyi Wan \\
  School of Architecture \\
  Carnegie Mellon University\\
  Pittsburgh, PA 15213 \\
  \texttt{linxiaow@andrew.cmu.edu} \\
}

\begin{document}

\maketitle


\section{Introduction and literature review}
New drug development is one of the most significant scientific challenge related to human welfare. Drugs work in vivo by interacting with bio-molecules in cells especially proteins. The structures of protein and their dynamic functions are the key information to design innovative molecules as drugs. However, until now, the number of U.S. Food and Drug Administration (FDA) approved drugs is 2715 \cite {jumper2021highly}. It is challenging to design new drug-like molecules based on current drugs because of the limited knowledge toward drug structures and drug-protein interactions. Additionally, Alpha-Fold \cite{wishart2018drugbank}, a deep learning algorithm for predicting protein structures, has extended our knowledge of protein structures from 198, 848 to 1,198,209 \cite{bernstein1977protein}. To generate more drug molecules based on those protein structures, deep-learning based computational tools are applied.

\begin{wrapfigure}{l}{0.25\textwidth} 
    \centering
    \caption{liGAN}
    \includegraphics[scale=0.5]{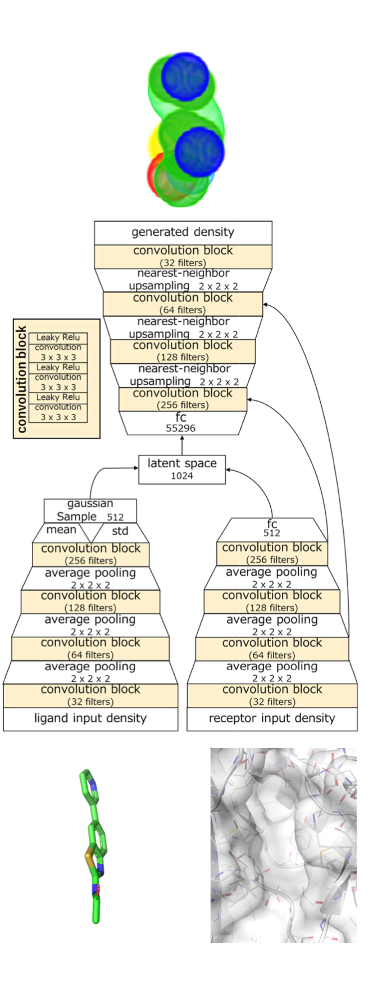}
\end{wrapfigure}

To generate new objects based on the known probability distribution,  a statistical model of the joint probability distribution between observed variable $X$ and targeted variable $Y$, named generative model, is developed so that we can get $Y$ given any observation $X = x$ as discriminate model. To learn the distribution of variables, two major families perform outstandingly and deserving more special attentions: Generative Adversarial Network (GAN) \cite{goodfellow2020generative} and Variational Autoencoders (VAE) \cite{cinelli2021variational}. GAN contains two neural networks named generator and discriminator respectively that the competition to each other perform as a zero-sum game: any gain in one network causes lose in other network. VAE, briefly speaking, is to learn current features $X$ from observed variables as "new features", which we call those features as latent space. We consider it as encoder process. Based on those "new features", we can generate target variables $Y$. We call it decoder process. Recently, an alternative method named flow-based generative model has been developed. Different from GANs, such algorithm learns likelihood function from observations and also explore the latent variables from the given sample. Different from GAN that has two neural networks and VAE that leaned latent variables, flow-based generative model, starting from very simple distribution (like Gaussian distribution), learned series of functions to update the simple distribution so that it is close to the probability distribution of observation data. 

To generate molecules from current protein-drug complexes, ligand Generative Adversarial Network (liGAN) \cite{masuda2020generating} uses GAN to generate molecules from current protein-drug complexes. It converts the coordinates of atoms from drugs and proteins as grid densities related to the type of atoms. The model contains two neural networks (Figure. 1). VAE model is also applied (Figure. 2)\cite{ragoza2022generating} to generate new models. Liu et.al. \cite{liu2022generating} demonstrated high validity and good binding affinity of the generated molecules using a flow-based generative model and a graph neural network (GNN). High validity means the generated small molecules are real chemicals rather than random atom groups while high binding affinity means the generated molecules have good shapes fitting the protein and also have good physio-chemical properties to interact with the protein. 

\begin{wrapfigure}{l}{0.5\textwidth} 
    \centering
    \caption{CVAE}
    \includegraphics[scale=0.13]{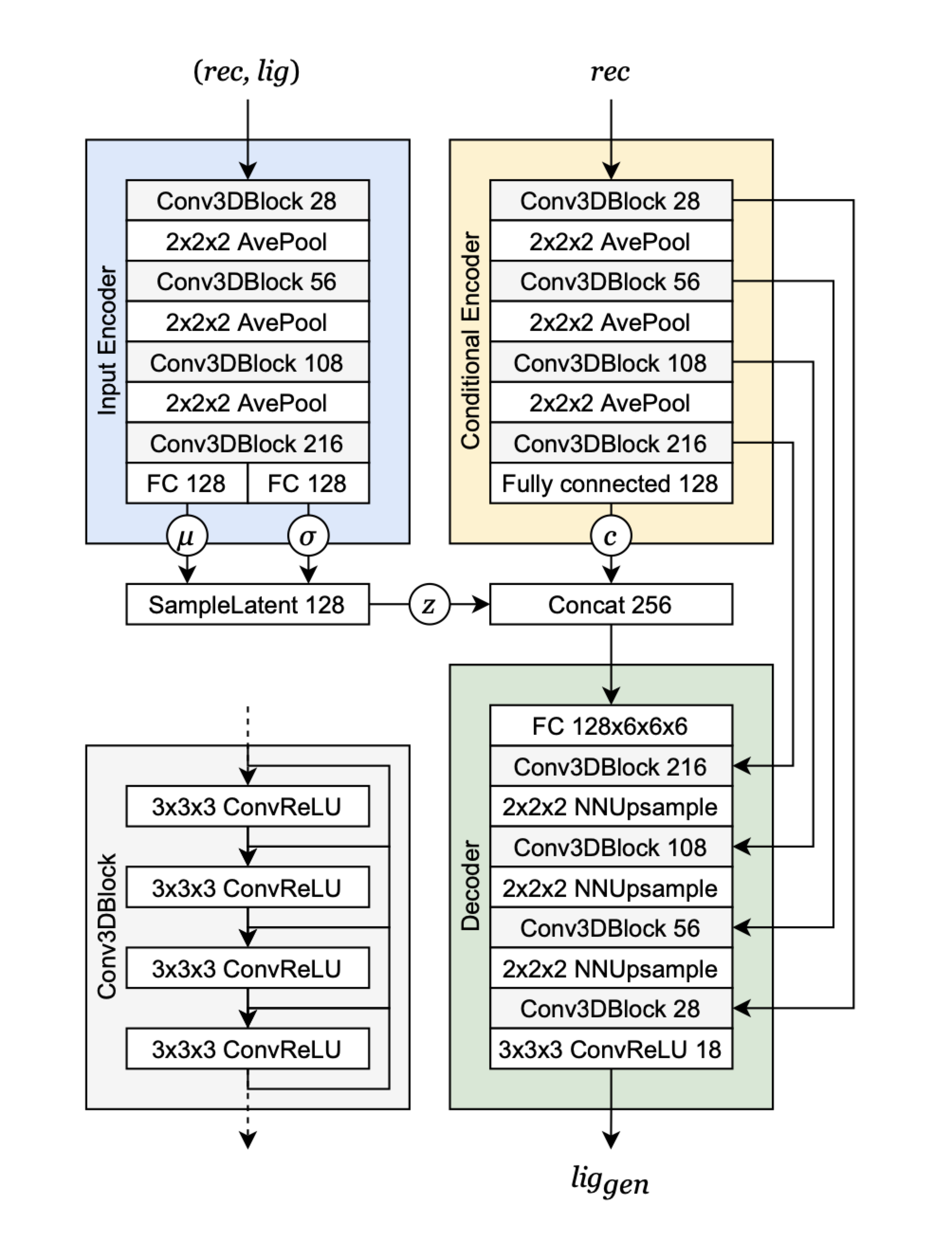}
\end{wrapfigure}

However, all those methods considered proteins as rigid body, refusing any change of structures, and only include very small part of proteins around binding site. However, dynamics of proteins are essential for both biological functions and drugs binding. The shape of binding pockets is changing because of the approaches of ligands. Additionally, local binding pocket cannot demonstrate the vicinities, which are defined as whether or not a binding pocket is approachable for small chemicals. Additionally, the training data people used are based on docking. The docking results are not always reliable because any given protein and small molecules can be docked with extreme twisted structures but it never happened in real condition.\cite{rayamit}

Starting from GraphBP model \cite{jumper2021highly}, we will use more solid training sets obtaining from Protein Data Bank (PDB) \cite{bernstein1977protein} and add extra context information for whole proteins. The results will be evaluated with validity and binding affinity by using rdkit \cite{landrum} package.

\section{Model description}
\label{gen_inst}

\subsection{Annotations}
To formalize the problem, the observation of drug-protein complex is annotated as both protein and drug atom types and coordinates. For drug information: $$M = {(a_i, r_i)}^n_\{i = 1\}$$  
In this equation, $n$ atoms included, $a_i$ is a vector represent atom type for $i$th atom and $r_i$ is its corresponded 3D Cartesian coordinate. 
For protein information: 
$$P = {(b_j, s_j)}^m_\{j = 1\}$$
$m$ atoms in protein binding pocket included, $b_j$ is atom type for $j$th atom and $s_j$ is its corresponded 3D Cartesian coordinate. 
The model is to learn $$p(M|P)$$
The generation drug atoms is also related to the previous drug atoms that has already been generated, which is annotated as $C^{(t)}$. $$C^{(t)} = P \cup \{(a_i, r_i) \}^{t}_{i = 1}$$

\subsection{Baseline Selection}
The baseline we selected to compare our model is liGAN \cite{masuda2020generating}. It is a very ideal generative adversarial network established to generate models. After generating new molecules, they used root mean square distance (RMSD) as metric to calculate the distance between original ligand and genertive ligand. RMSD defines as the average of square root of atom-atom distances between two molecules, indicating how similar they are structurally. In their work, RMSDs between the reference drug and generative molecule were calculated to evaluate the similarities between real molecules and generative molecules. Additionally, the binding affinities of molecules they generated are 4 to 8, which is acceptable as leading compounds. However, the validity they did not report. The flow-based generative model we used also reported the binding affinity outperformed liGAN, showing much better binding affinities compared with the reference molecules.  

\subsection{Flow-based generative model}
\begin{figure}[h]
\caption{flow-based model}
\centering
\includegraphics[scale=0.5]{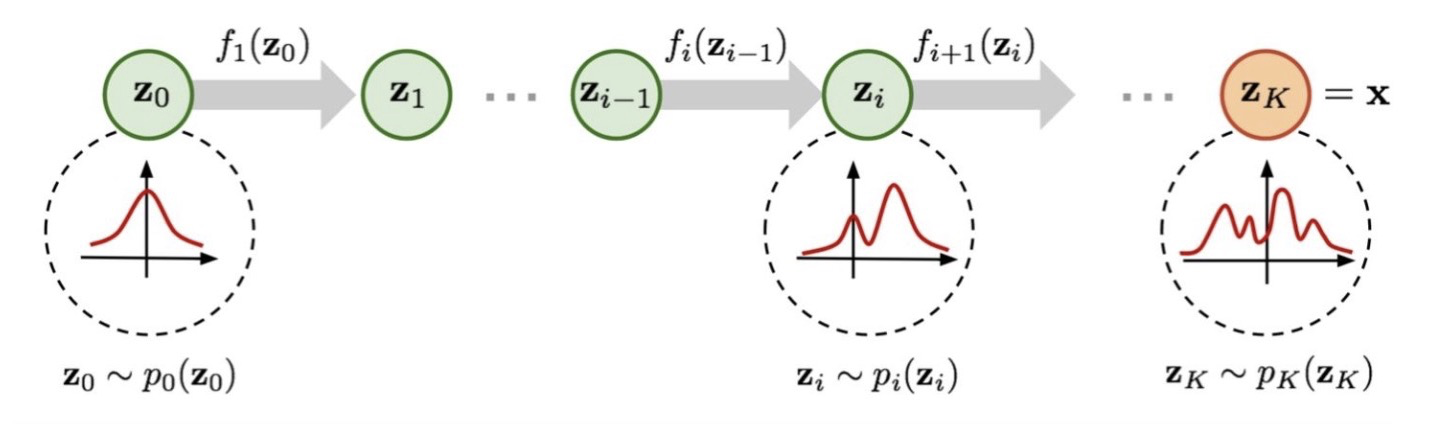}
\end{figure}
The basic idea for flow-based generative model is to learn a series of functions such that the distribution of observed data can be simulated from simple distributions $z_0$ like Gaussian distribution or uniform distribution. The distribution of observed data $x$ can be represented by: $$x = z_K = f_K \circ f_{K - 1} \circ f_{K - 2} \circ ... \circ f_i \circ ... \circ f_2 \circ f_1(z_0)$$
For step $i$, the distribution $z_i$ and function $f_i$ are determined by the previous distribution $z_{i - 1}$, we have
$$p_i (z_i) = p_{i - 1} (f^{-1}_i (z_i)) |{det \frac{df_i^{-1}}{dz_i}}| $$
The $det$ is the determinant of matrix $\frac{df_i^{-1}}{dz_i}$. To train the model, function $f$ should be reversible and its Jocobian matrix is required to be positive semi-definite so that we can directly use the diagonal of matrix $f_i$ to get the results of determinant. Based on the inverse function theorem and the properties of Jocobians, we have $$log p_i (z_i) = log p_{i-1} (z_{i-1}) - log |det \frac{df_i}{dz_{i-1}}|$$
From the previous results, the distribution of $x$ is:
$$log p(x) = log \pi_K (Z_K) = log \pi_{K-1} (z_{K -1}) - log |det \frac{df_K}{dz_{K - 1}}| = log \pi_0 (z_0) - \sum _{i=1}^K log |det \frac{df_i}{dz_{i - 1}}|$$
To train the model, we just maximize the log-likelihood:
$$Loss = -log p(x)$$

Specifically, to generate a drug, atom types of drugs are generated by $$a_t = g^a (C^{(t-1)}; z^a_t)$$
Corresponded coordinates are generated by
$$r_t = g^r (C^{(t-1)}, a_t; z^r_t)$$
The $z^a_t$ and $z^r_t$ are latent variables. The context is updated by 
$$C^{(t)} = C^{(t - 1)} \cup \{(a_t, r_t)\}$$

The function from flow-based model $g$ we learned is $g^a(\cdot)$ and $g^r(\cdot)$

\subsection{Graph neural network}
Graph neural network \cite{scarselli2008graph} (GNN) is based on data structure graph. It contains vertices (V) and edges (E). $$G = (V, E)$$ As a neural network, it can learn the information of node-level, edge-level and graph-level predictions. In this section, the features of atoms including atom type and coordinate can be learned.
$$V = \{(a, b)\}$$
The vertices $V$ indicate atom types from drug $a$ and from protein $b$.
$$E = \{(r, s)\}$$
The edges $E$ indicate drug atom coordinates $r$ and protein atom coordinates $s$.
For $k$th layer of the network, we have
$$Z_{V, E}^k = \sigma (W_k \sum \frac{Z_{V, E}^{k-1}}{N(V, E)} + B_k Z_{V, E}^k)$$
$W_k$ is the weight for GNN, $B_K$ is the bias, and $\sigma$ is any non-linearity activation.

In our model, for $l$th layer GNN at time $t$, the value $h_k^{(t, l)}$ determines atom k in context with $C^{(t)}$, we have
$$h_k^{(t, l)} = h_k^{(t, l-1)} + \sum_{u\in N(k)} h_u^{(t, l-1)} \odot MLP^l (e_{RBF}(d_{uk}))$$
MLP is multilayer perceptron, RBF is radial basis function kernel defined as $$K(x, x') = e^{-\frac{||x - x'||^2}{2 \sigma^2}}$$
$\sigma$ in this equation is a free parameter.

\subsection{Evaluation metric}
Validity implies if there exists any illegal atoms or illegal chemical bonds. For example, carbon can only generate 4 chemical bonds, nitrogen can generate 3 covalent bonds and one ionic bond, and oxygen can formalize 2 bonds. Any chemical is considered as illegal if it violates the basic chemistry rules. The SMILES chemical notation language is to convert chemicals to a string by using fingerprint sampling algorithms. By using "Chem" package, we can learn if those chemicals we generated are the real molecules.

Binding affinity elucidates whether or not the generative molecules have good shapes and physio-chemical properties to bind the given protein. The higher binding affinity indicates the better pharmacological effects for the further treatment effects. PRODIGY (PROtein binDIng enerGY prediction) \cite{xue2016prodigy} is a solid machine learning tools to predict binding affinities based on charged/uncharged interactions, hydrophobic/hydrophilic properties, and hydrogen bonds to predict the dissociation constants $K_d$:
$$\Delta G = RT ln K_d$$
Where $\Delta G$ is the binding affinity, $R$ is the idea gas constant, and $T$ is temperature.


\section{Proposed extension}
To train better model, scPDB \cite{desaphy2015sc} is applied. Different from CrossDocked2020 data. It includes 16034 entries, 4782 proteins and 6326 ligands. For each entry, it contains one or more binding complexes. In total, it contains more than 50,000 drug-protein binding structures. Compared with docking results from CrossDocked2020, all binding structures are from experiments rather than docking predictions. Additionally, the dynamics of proteins can be implemented by using attention based graphical model. The atoms are emphasized which have higher B-factors. The B-factor describes the displacement of atomic position from mean-square displacement. High B-factor for one atom indicates the atom is more flexible and vice versa. Considering the vibrations of atoms, attention based graphical models are applied to include B-factors for each atoms. 




\medskip

\printbibliography

\end{document}